\documentclass[letterpaper]{article} % DO NOT CHANGE THIS
\usepackage[draft]{aaai25}  % DO NOT CHANGE THIS
\usepackage{times}  % DO NOT CHANGE THIS
\usepackage{helvet}  % DO NOT CHANGE THIS
\usepackage{courier}  % DO NOT CHANGE THIS
\usepackage[hyphens]{url}  % DO NOT CHANGE THIS
\usepackage{graphicx} % DO NOT CHANGE THIS
\usepackage[numbers]{natbib}
\usepackage{caption} % DO NOT CHANGE THIS AND DO NOT ADD ANY OPTIONS TO IT
\usepackage{multirow} %dy
\usepackage{rotating} %dy
\usepackage[table,xcdraw]{xcolor} %dy
\usepackage{colortbl} %dy
\usepackage{array} %dy
\usepackage{tikz} %dy
\usepackage{adjustbox} %dy
\usepackage{tabularx} %dy
\usepackage{booktabs} %dy
\usepackage{amssymb} %dy
\usepackage{booktabs} %cw
\usepackage{placeins}
\usepackage{xcolor}
\usepackage{hyperref}
\usepackage{titlesec} % titlesec 패키지 사용

\hypersetup{
    colorlinks=true,
    citecolor=green,
    linkcolor=red,
    urlcolor=magenta,
}

\usepackage{algorithm}
\usepackage{algorithmic}
\usepackage{pifont}%dy
\usepackage{newfloat}
\usepackage{listings}
\DeclareCaptionStyle{ruled}{labelfont=normalfont,labelsep=colon,strut=off} % DO NOT CHANGE THIS
\floatstyle{ruled}
\newfloat{listing}{tb}{lst}{}
\floatname{listing}{Listing}
\usepackage{amsfonts} %cw addition
\usepackage{amsmath} %cw addition

\title{ProtoOcc: Accurate, Efficient 3D Occupancy Prediction Using \\ Dual Branch Encoder-Prototype Query Decoder
}
\author{
    Jungho Kim\textsuperscript{\rm 1}\equalcontrib \quad 
    Changwon Kang\textsuperscript{\rm 2}\equalcontrib \quad 
    Dongyoung Lee\textsuperscript{\rm 2}\equalcontrib \quad 
    Sehwan Choi\textsuperscript{\rm 2} \quad 
    Jun Won Choi\textsuperscript{\rm 1}\thanks{Corresponding author}
}
\affiliations{
    \textsuperscript{\rm 1}Seoul National University \hspace{0.5cm}
    \textsuperscript{\rm 2}Hanyang University\\ 
}

\begin{document}

\maketitle

\begin{abstract}

In this paper, we introduce ProtoOcc, a novel 3D occupancy prediction model designed to predict the occupancy states and semantic classes of 3D voxels via a deep semantic understanding of scenes. ProtoOcc consists of two main components: the \textit{Dual Branch Encoder} (DBE) and the \textit{Prototype Query Decoder} (PQD). The DBE produces a new 3D voxel representation by combining 3D voxel and BEV representations across multiple scales using a dual branch structure. This design combines the BEV representation, which offers a large receptive field, with the voxel representation, known for its higher spatial resolution, thereby improving both performance and computational efficiency.
The PQD employs two types of prototype-based queries to expedite the Transformer decoding process. Scene-Adaptive Prototypes are generated from the 3D voxel features of the input sample, while Scene-Agnostic Prototypes are updated during training using an Exponential Moving Average of the Scene-Adaptive Prototypes.
Using these prototype-based queries for decoding, we can directly predict 3D occupancy in a single step, eliminating the need for iterative Transformer decoding.
Additionally, we propose \textit{Robust Prototype Learning}, which introduces noise into the prototype generation process and trains the model to denoise during the training phase. This approach enhances the robustness of ProtoOcc against degraded prototype feature quality. ProtoOcc achieves state-of-the-art performance with 45.02\% \textit{mIoU} on the Occ3D-nuScenes benchmark. For the single-frame method, it reaches 39.56\% \textit{mIoU} with 12.83 FPS on an NVIDIA RTX 3090. Our code can be found at \href{https://github.com/SPA-junghokim/ProtoOcc}{https://github.com/SPA-junghokim/ProtoOcc}.

\end{abstract}

\setcounter{secnumdepth}{2}
\renewcommand{\thesection}{\arabic{section}.} % 섹션 번호 뒤에 점 추가
\renewcommand{\thesubsection}{\thesection\arabic{subsection}.} % 서브섹션

\titlespacing*{\section}{0pt}{1.2em}{0.5em} % 섹션 위 간격 1.5em, 아래 간격 1em
\titlespacing*{\subsection}{0pt}{1.0em}{0.55em} % 서브섹션 위 간격 1em, 아래 간격 0.5em

\titleformat{\section}{\normalfont\Large\bfseries\centering}{\thesection}{0.5em}{}
 \titleformat{\subsection}{\normalfont\large\bfseries}{\thesubsection}{0.5em}{}

%\titlespacing*{\subsubsection}{0pt}{0.63em}{0.05em} % 서브서브섹션 위 간격 0.75em, 아래 간격 0.5em

%\titleformat{\section}{\normalfont\Large\bfseries}{\thesection}{0.5em}{}
% \titleformat{\subsubsection}{\normalfont\normalsize\bfseries}{\thesubsubsection.}{0.5em}{}

%\titleformat{\subsubsection}[runin]{\normalfont\normalsize\bfseries}{\thesubsubsection.}{0.5em}{}[\hspace{1em}]

% \titleformat{\subsubsection}[runin]{\normalfont\normalsize\bfseries}{\thesubsubsection.}{10em}{} % ":" 제거 후 간격 추가

\section{Introduction}
Vision-based 3D occupancy prediction is a critical task for comprehensive scene understanding around the ego vehicle in autonomous driving. This task aims to simultaneously estimate occupancy states and semantic classes using multi-view images in 3D space, providing detailed 3D scene information. The typical prediction pipeline of previous methods comprises three main components: 1) a view transformation module, 2) an encoder, and 3) a decoder.
Initially, backbone feature maps extracted from multi-view images are transformed into 3D spatial representations through a 2D-to-3D view transformation. An encoder network then processes these 3D representations to produce high-level semantic spatial features, capturing the overall scene context. Finally, a decoder network utilizes these encoded 3D spatial features to predict both semantic occupancy and class for all voxels composing the scene.

\begin{figure}[!t]
    \centering
    \includegraphics[scale=0.72]{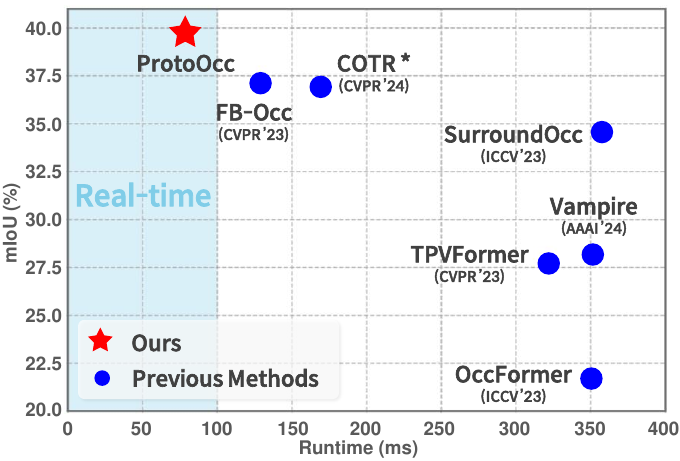}
    \caption{Comparisons of the \textit{mIoU} and runtimes of different methods on the Occ3D-nuScenes validation set. $\star$ indicates results reproduced using publicly available codes. Inference time is measured on a single NVIDIA RTX 3090 GPU.  
    }
    \label{intro_1}
\end{figure}

\begin{figure*}[t!]
    \centering
    \includegraphics[scale=0.39]{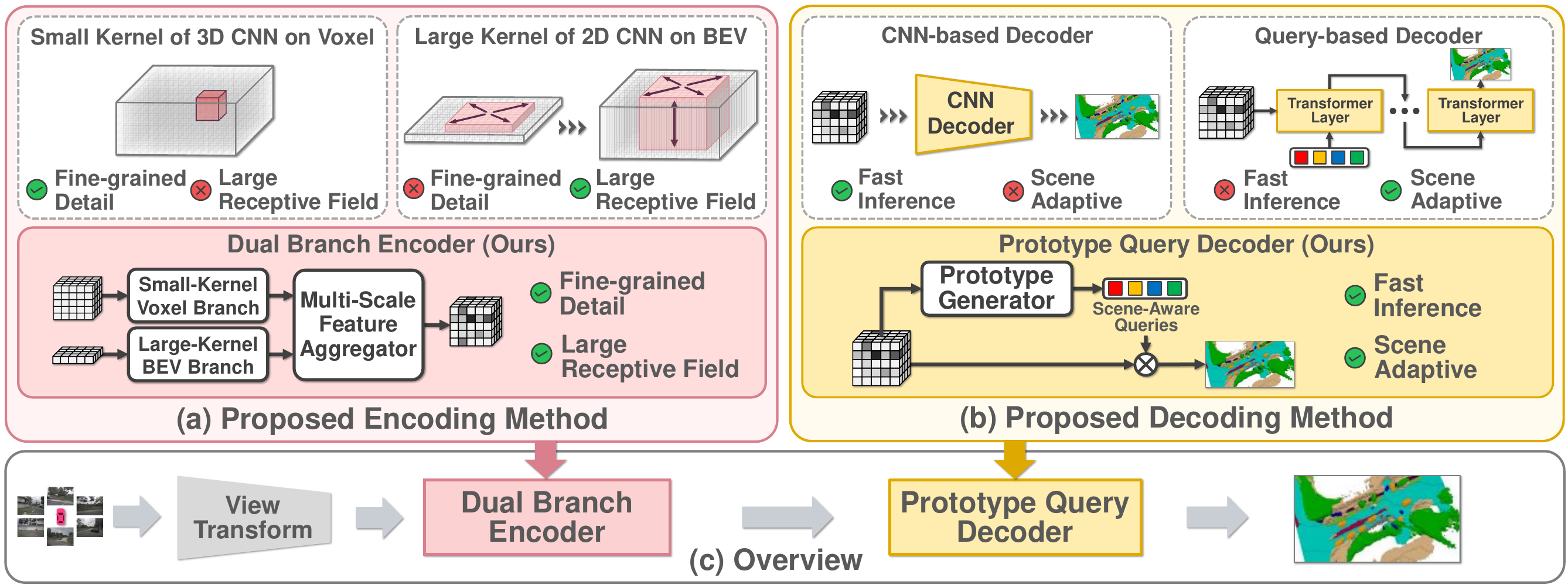}
    \caption{
    Overall structure of ProtoOcc.
    (a) Dual Branch Encoder captures fine-grained 3D structures and models the large receptive fields in voxel and BEV domains, respectively. 
    (b) The Prototype Query Decoder generates Scene-Aware Queries utilizing prototypes and achieves fast inference without iterative query decoding. (c) Our ProtoOcc framework integrates Dual Branch Encoder and Prototype Mask Decoder for 3D occupancy prediction. 
    }
    \label{intro_2}
\end{figure*}

Existing works have explored enhancing encoder-decoder networks to improve both the accuracy and computational efficiency of 3D occupancy prediction. Various attempts have been made to optimize encoders using 3D spatial representations. Figure \ref{intro_2} (a) illustrates two commonly used 3D representations, including voxel representation \cite{VoxFormer, Openoccupancy, Surroundocc} and Bird's-Eye View (BEV) representation \cite{FastOcc, flashocc}.
Voxel-based encoding methods \cite{OccFormer, Monoscene} used 3D Convolutional Neural Networks (CNNs) to encode voxel structures. However, the large number of voxels needed to represent 3D surroundings results in high memory and computational demands. While reducing the capacity of 3D CNNs can alleviate this complexity, it also reduces the receptive field, which may compromise overall performance.

Unlike voxel representations, BEV representations project 3D information onto a 2D BEV plane, significantly reducing memory and computational requirements. After encoding the BEV representation using 2D CNNs, it is converted back into a 3D voxel structure for 3D occupancy prediction. However, this approach inherently loses detailed 3D geometric information due to the compression of the height dimension. Although incorporating additional 3D information \cite{FastOcc, flashocc} can enhance BEV representation, its performance remains limited by the inherent constraints of representing 3D scenes in a 2D format.

Another line of research focuses on enhancing the decoders. As illustrated in Figure \ref{intro_2} (b), two main decoding strategies exist: 1) CNN-based decoders \cite{Monoscene, sogdet, Vampire, RadOcc} and 2) query-based decoders \cite{OccFormer, SparseOcc_Rethinking, SparseOcc_Fully}.
CNN-based decoders employed lightweight 3D CNNs to extract semantic voxel features, while query-based decoders iteratively decoded a query using the 3D representation obtained from the encoder. 
Although query-based decoders achieved better prediction accuracy, they required processing through multiple decoding layers, leading to increased inference time. Therefore, it is crucial to reduce this complexity while retaining the performance benefits of query-based decoders.

%  for 'a' 3D occupancy prediction (Grammarly)
To address the aforementioned challenges, we introduce ProtoOcc, an efficient encoder-decoder framework for a 3D occupancy prediction network. 
As shown in Figure \ref{intro_1}, ProtoOcc achieves state-of-the-art performance while achieving relatively fast inference (i.e., 77.9 ms) on a single NVIDIA RTX 3090 GPU.
%Additionally, when higher computational complexity is permissible, ProtoOcc achieves performance among current 3D occupancy prediction methods. 

As shown in Figure \ref{intro_2} (a), ProtoOcc utilizes a \textit{Dual Branch Encoder} (DBE) with a dual-branch architecture.
The voxel branch uses 3D CNNs with small kernel sizes to reduce computational complexity, while the BEV branch applies 2D CNNs with large kernel sizes to capture scene semantics with a larger receptive field. To combine the strengths of both representations, BEV and voxel features are fused across multiple scales to generate {\it Comprehensive Voxel Feature}. This dual encoding approach effectively captures fine-grained 3D structures and long-range spatial relationships across various scales.

Query-based decoding typically demands high computational complexity due to processing across multiple decoding layers. To overcome this, we propose the \textit{Prototype Query Decoder} (PQD), which accelerates the decoding process by utilizing prototype-based queries and eliminating the need for iterative decoding.
PQD generates Scene-Adaptive Prototypes by utilizing class-specific masks to aggregate features for each class from the Comprehensive Voxel Feature.
While these prototypes can represent the semantic classes present in the input, challenges arise when certain semantic classes are absent in the input sample. To address this, we introduce Scene-Agnostic Prototypes, which are generated by accumulating Scene-Adaptive Prototypes across samples using an Exponential Moving Average (EMA) during training. By combining Scene-Adaptive and Scene-Agnostic Prototypes together, PQD forms Scene-Aware Queries, enabling efficient 3D occupancy prediction in a single iteration.

%This approach enables us to complete the 3D occupancy prediction task quickly without the need for an iterative decoding process.

We also develop a novel training method for enhancing the performance of the proposed decoder.
Since the prototypes are directly utilized for 3D occupancy prediction without an iterative query decoding, the quality of the prototypes significantly impacts the overall performance.
To ensure robust predictions, we devise the \textit{Robust Prototype Learning} framework that injects noise into the prototype generation process and trains the model to counteract this noise during the training phase.

We evaluated ProtoOcc on the challenging Occ-3D nuScenes benchmark \cite{Occ3d}. ProtoOcc achieves an \textit{mIoU} of $\mathbf{39.56}$\%, surpassing the performance of all existing single-frame methods, while operating at a processing speed of $\mathbf{12.83}$ FPS on an NVIDIA RTX 3090.
Combined with multi-frame temporal fusion, ProtoOcc also achieves state-of-the-art performance among the latest multi-frame methods, with an \textit{mIoU} of $\mathbf{45.02}$\%.

The contributions of this study are summarized below:
\begin{itemize}

\item We introduce ProtoOcc, a novel 3D occupancy prediction model that integrates a dual-branch encoding and query-based decoding to enhance both computational efficiency and accuracy for complex 3D environments.

\item We propose an enhanced 3D representation for the encoder that jointly aggregates voxel and BEV representations through dual branch pipelines. This DBE method efficiently allocates resources, forming the largest receptive field with minimal computational cost.

\item We propose a computationally efficient decoder performing 3D occupancy prediction in a single pass. This PQD generates queries representing each class from the encoded 3D spatial features and directly predicts semantic occupancy without a decoding process, thereby significantly reducing the computational complexity.

\item ProtoOcc achieves state-of-the-art performance, with a 45.02\% \textit{mIoU} on the Occ-3D benchmark. It also achieves a 39.56\% \textit{mIoU} at a processing speed of 12.83 FPS.

% \item The codes will be publicly available. 

\end{itemize}

\section{Related Works}
\subsection{3D Encoding Methods for Occupancy Prediction}
3D occupancy prediction \cite{OccNet} has attracted considerable interest in recent years due to its ability to reconstruct 3D volumetric scene structures from multi-view images.
These approaches primarily utilize two widely adopted 3D representations, voxel and BEV, to encode 3D spatial information.
MonoScene \cite{Monoscene} bridged the gap between 2D and 3D representations by projecting 2D features along their line of sight and encoding voxelized semantic scenes with a 3D UNet.
OccFormer \cite{OccFormer} introduced a dual-path transformer that independently processes voxel and BEV representations, dividing voxel data into BEV slices to decompose heavy 3D processing.
FastOcc \cite{FastOcc} reduced computational cost by replacing high-cost 3D CNNs in voxel space with efficient 2D CNNs in BEV space.

\subsection{3D Decoding Methods for Occupancy Prediction}
Recent studies \cite{lowrankocc, PaSCo, SparseOcc_Fully, SparseOcc_Rethinking} have introduced query-based decoders that capture scene-adaptive features by interacting with voxel features.
OccFormer \cite{OccFormer} adopted masked attention in 3D space to iteratively decode query embeddings, thereby extracting semantic information from voxel features.
COTR \cite{COTR} introduced a coarse-to-fine semantic grouping strategy, dividing categories into semantic groups based on granularity and assigning distinct supervision for each group to address class imbalance.

\subsection{2D Encoding Methods with Large Receptive Fields}
Transformer-based models, such as ViT \cite{vits} and Swin Transformer \cite{Swin}, have gained significant popularity in the field of computer vision.  
Recent studies \cite{ERF, ConTNet} have shown that large receptive fields are a crucial factor in the success of these models.
Recent research on CNN-based models has demonstrated that models with large receptive fields can achieve competitive performance with Transformer-based architectures.
% Recent research has demonstrated that CNN-based model with large receptive fields can achieve competitive performance with Transformer-based architectures.
ConvNeXt \cite{ConvNeXt} achieved competitive performance by modifying ConvNets with design principles from vision Transformers, including 7×7 depth-wise convolutions.
RepLKNet \cite{RepLKNet} scaled up convolutional kernels to as large as 31×31 utilizing re-parameterization. 
LargeKernel3D \cite{Largekernel3d} proposed spatial-wise partition convolutions, achieving a large receptive field in 3D while reducing computational costs.

\section{ProtoOcc Method}
\subsection{Overview}
The overall architecture of ProtoOcc is illustrated in Figure \ref{intro_2} (c). Initially, a 2D-to-3D view transformation generates both 3D voxel and BEV features from multi-view camera images. DBE then combines these features across multiple scales to produce Comprehensive Voxel Feature. 
Next, PQD produces class-specific Scene-Aware Queries from the Comprehensive Voxel Feature and utilizes them to predict 3D occupancy in a single pass.

\subsubsection{2D-to-3D View Transformation.}
The 2D-to-3D View transformation process converts multi-view camera inputs into 3D features in both voxel and BEV formats through Lift-Splat-Shoot (LSS) method \cite{LSS}. 2D feature maps are extracted from multi-view images using a backbone network such as ResNet \cite{ResNet}. These features are then fed into a depth network to predict depth distributions. Frustum features are generated by computing the outer product between 2D feature maps and depth distributions. The voxel-pooling method transforms these frustum features into a unified 3D voxel feature $F_{\text{vox}}$. Finally, the BEV feature $F_{\text{BEV}}$ is reshaped from $F_{\text{vox}}$ along the Z axis, changing from $(D, X, Y, Z)$ to $(D$$\times$$Z, X, Y)$, where $D$ denotes the channel dimension and $(X, Y, Z)$ represents the volume scale.

\subsection{Dual Branch Encoder}
\begin{figure}[t]
    \centering
    \includegraphics[scale=0.35]{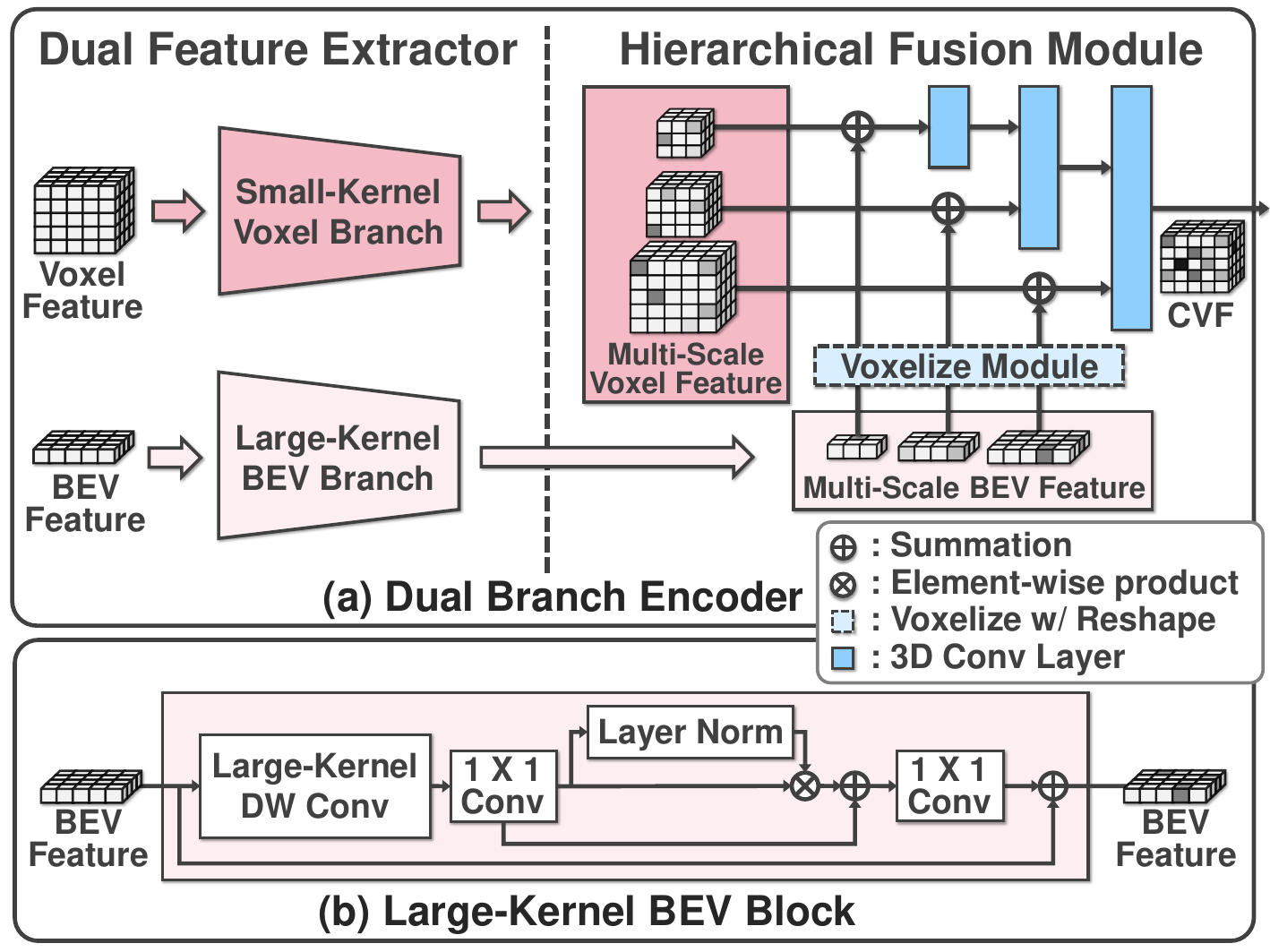}
    \caption{Details of Dual Branch Encoder. 
    (a) DBE consists of DFE and HFM. DFE extracts multi-scale features using the dual encoders in the voxel and BEV domain. 
    HFM aggregates these features from low to high scales to generate Comprehensive Voxel Feature $V_{\text{CVF}}$.
    (b) The Large-Kernel BEV Block comprises a large kernel depth-wise convolution, 1x1 convolutions, and layer normalization.
    }
    \label{vis_DBE}
\end{figure}

The structure of DBE is depicted in Figure \ref{vis_DBE} (a).
DBE consists of two main components: the \textit{Dual Feature Extractor} (DFE) and the \textit{Hierarchical Fusion Module} (HFM). 
The DFE module captures fine-grained 3D structures in the voxel domain and long-range spatial relationships in the BEV domain, extracting features across multiple scales. 
The HFM module hierarchically aggregates features from each domain, generating comprehensive context representations at various levels of detail.

\subsubsection{Dual Feature Extractor.}
DFE consists of a voxel branch with 3D CNNs and a BEV branch with 2D CNNs designed for distinct spatial representations.
The voxel branch aims to efficiently extract fine-grained features by utilizing small kernels to minimize computational complexity. 
\(F_{\text{vox}}\) is processed through 3D CNN residual blocks and downsampling layers, generating multi-scale voxel features \(\mathbf{V}^{\text{vox}} = \{ V^{\text{vox}}_i \in \mathbb{R}^{D_i \times X_i \times Y_i \times Z_i} \}_{i=1}^{S}\), where \(i\) denotes the scale index and \(S\) represents the total number of scales.

The BEV branch is designed to capture long-range spatial relationships by utilizing 2D CNNs with larger kernel sizes, which effectively expand the receptive field. This approach avoids the high computational burden required by 3D CNNs.
Multi-scale BEV features \(\mathbf{B}^{\text{BEV}} = \{ B^{\text{BEV}}_i \in \mathbb{R}^{D'_i \times X_i \times Y_i} \}_{i=1}^{S}\) are extracted from \(F_{\text{BEV}}\) through a series of 2D CNN residual blocks followed by a downsampling layer. 

\begin{figure}[t!]
    \centering
    \includegraphics[scale=0.35]{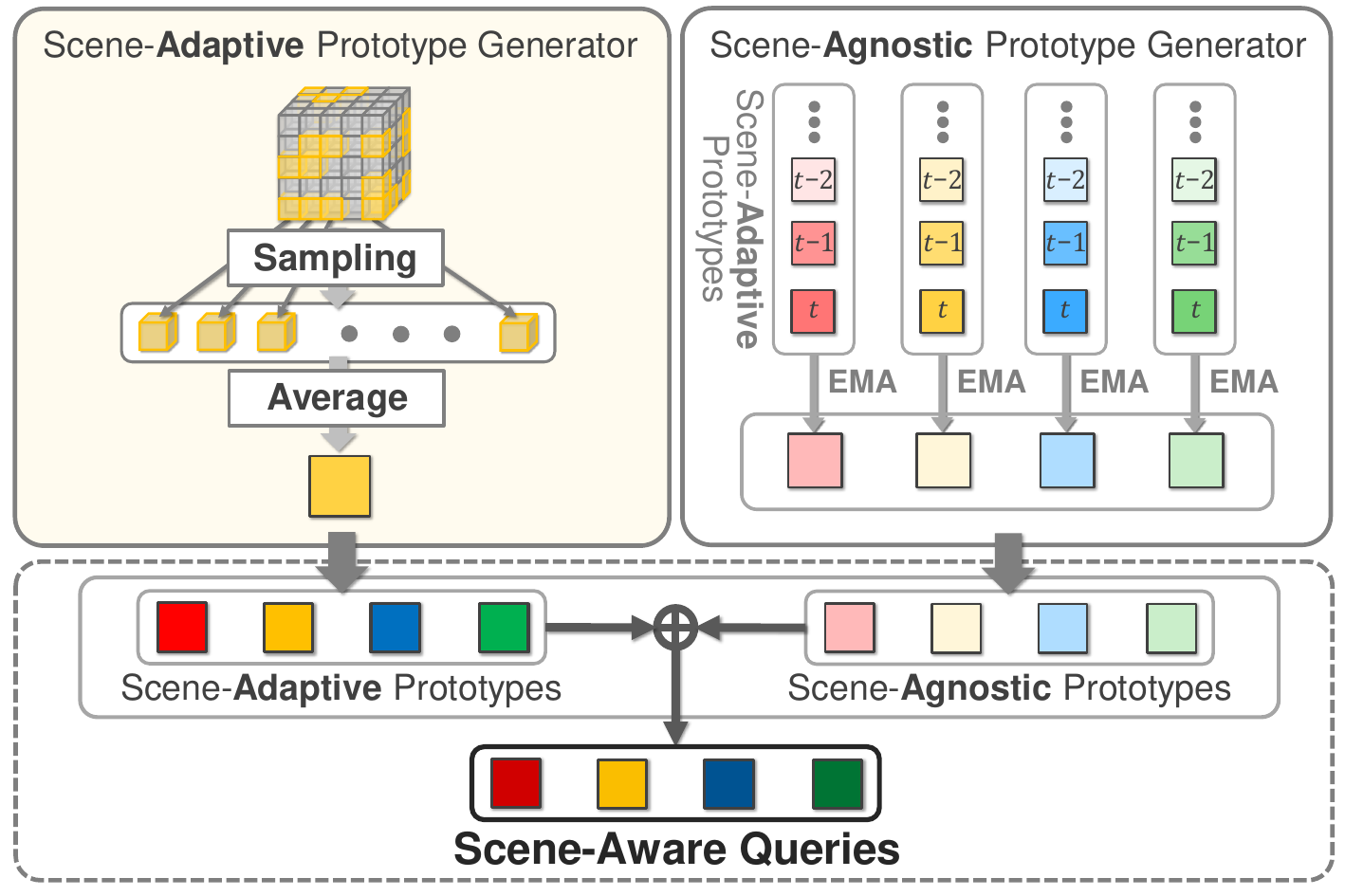}
    \caption{Details of prototype generation.
    AdaPG generates Scene-Adaptive Prototypes by sampling and averaging Comprehensive Voxel Feature for each class based on class-specific masks.
    AgnoPG generates Scene-Agnostic Prototypes by computing Scene-Adaptive Prototypes through the EMA method. Finally, Scene-Adaptive Prototypes and Scene-Agnostic Prototypes are combined into Scene-Adaptive Queries. 
    }
    \label{vis_QueryGnerator}
\end{figure}

\subsubsection{Hierarchical Fusion Module.}
HFM integrates multi-scale voxel and BEV representations to generate the Comprehensive Voxel Feature. This process involves hierarchical aggregation of features from both domains through a sequence of upsampling layers and a 3D CNN. 
In each layer, the BEV feature $B^{\text{BEV}}_i$ at the $i$-th scale is voxelized into $V^{\text{BEV}}_i \in \mathbb{R}^{D_i \times X_i \times Y_i \times Z_i}$ through a reshape operation, aligning it with the voxel feature space.
Subsequently, the fused voxel feature $V^{\text{fused}}_i$ is derived by combining the voxel feature $V^{\text{vox}}_i$, the voxelized BEV feature $V^{\text{BEV}}_i$, and the upsampled fused voxel feature $Up(V^{\text{fused}}_{i-1})$ from the previous layer as follows
\begingroup
\fontsize{9.5}{11.5}\selectfont
\begin{equation}
V^{\text{fused}}_i = 
\begin{cases}  
Conv (Up(V^{\text{fused}}_{i-1}) + V^{BEV}_i + V^{\text{vox}}_i) & \text{for } i > 1 \\
Conv (V^{BEV}_i + V^{\text{vox}}_i) & \text{for } i = 1
\end{cases},
\end{equation}
where $Up$ denotes upsampling layer by trilinear interpolation and ${Conv}$ denotes a 3D convolution layer with a small kernel size.
After processing through $S$ upsampling layers, DBE ends up with Comprehensive Voxel Feature $V_{\text{CVF}}$ $=$ $V^{\text{fused}}_{S}$.

\subsection{Prototype Query Decoder}
As illustrated in Figure  \ref{vis_QueryGnerator}, PQD comprises two components: the \textit{Scene-Adaptive Prototype Generator} (AdaPG) and the \textit{Scene-Agnostic Prototype Generator} (AgnoPG). The AdaPG generates Scene-Adaptive Prototypes to capture the unique features of each class in the current scene. AgnoPG produces Scene-Agnostic Prototypes across diverse scenes using the EMA method \cite{EMA}, mitigating challenges arising from missing certain classes and capturing comprehensive features for each class.
Finally, PQD predicts semantic occupancy for all voxels through a single step operation that leverages the Comprehensive Voxel Feature and the prototype-based queries.

\subsubsection{Scene-Adaptive Prototype Generator.}
AdaPG aims to generate Scene-Adaptive Prototypes that encapsulate class-specific features extracted from Comprehensive Voxel Feature of the current scene.
First, the AdaPG uses a shallow 3D CNN classifier to produce voxel-wise class probabilities $O_s \in \mathbb{R}^{C \times X \times Y \times Z}$, where $C$ denotes the number of semantic categories, including the empty class. These probabilities are utilized to construct class-specific binary masks $M^{cls}_{c}$ for each class \(c \in \{1,\dots,C\}\), as follows
\begin{equation}
M^{cls}_{c}(x,y,z) =
\begin{cases}
1 & \text{if } \underset{\tilde{c} \in \{1,\dots,C\}}{\mathrm{argmax}} \; O_s(x, y, z) = c  \\
0 & \text{otherwise}
\end{cases}.
\end{equation}
The resulting $M^{cls}_{c}$ is used to sample the voxel features for the $c$-th class. Subsequently, the Scene-Adaptive Prototypes $\mathbf{P}^{d}$$=$$\{P^{d}_{c} \in \mathbb{R}^{D}\}^{C}_{c=1}$ are derived by aggregating the sampled voxel features for each class through average pooling in both $x$, $y$, and $z$ domains
\begin{equation}
P^{d}_{c} = \frac{1}{N^{nz}_{c}} \sum_{(x,y,z)} (M^{cls}_{c}(x,y,z) \otimes V_{\text{CVF}}(x,y,z)),
\end{equation}
where $N^{nz}_{c}$ denotes the number of non-zero voxels in $M^{cls}_{c}$ and $\otimes$ is the element-wise product. When $N^{nz}_{c}$ is zero, $P^{d}_{c}$ is set to a zero vector.
The resulting $\mathbf{P}^{d}$ are delivered to the AgnoPG for query generation process.

\subsubsection{Scene-Agnostic Prototype Generator.}
While AdaPG effectively captures class-specific features within the current scene, the absence of sampled features for certain classes results in incomplete prototypes.
To address this, the AgnoPG generates Scene-Agnostic Prototypes $\mathbf{P}^{g}$ by applying the EMA \cite{EMA} method to $\mathbf{P}^{d}$, continuously integrating features across diverse scenes. That is, for each iteration, $\mathbf{P}^{g}$ is updated as
\begin{equation}
\mathbf{P}^{g}(t) = \alpha \cdot \mathbf{P}^{d}(t) + (1-\alpha) \cdot \mathbf{P}^{g}(t-1),
\end{equation}
where $t$ denotes the iteration index and $\alpha$ is the EMA coefficient. 
This process ensures the generation of comprehensive prototype features encompassing all classes.

\subsubsection{Prototype-Driven Occupancy Prediction.}

Scene-Aware Queries $Q^{SA} \in \mathbb{R}^{C \times D}$ are generated by combining $\mathbf{P}^{d}$ from AdaPG and $\mathbf{P}^{g}$ from AgnoPG through summation.  
Notably, the occupancy prediction results are obtained directly from the Scene-Aware Queries, eliminating the need for iterative Transformer decoding.
The Scene-Aware Queires are processed through MLP layers to predict semantic logits $p_c$ and mask embedding $\varepsilon^{\text{mask}}_{c}$ for each class  $c$. 
Subsequently, the occupancy masks $M^{\text{occ}}_{c}$ are generated by performing a dot product between the Comprehensive Voxel Feature and the mask $\varepsilon^{\text{mask}}_{c}$
  along the channel dimension, followed by the application of a sigmoid function to normalize the resulting masks.
Finally, the 3D semantic occupancy prediction $\mathbf{O}_{s}$ is obtained
\begin{equation}
\mathbf{O}_{s} = \sum^{C}_{c=1} p_c \cdot M^{\text{occ}}_{c}.
\end{equation}
Our approach simplifies the decoding process by processing prototype-based queries in a single step.

\newcommand{\coloredsquare}[1]{\textcolor{#1}{\rule{2.5mm}{2.5mm}}}

\definecolor{others}{RGB}{0, 0, 0}
\definecolor{barrier}{RGB}{112, 128, 144}
\definecolor{bicycle}{RGB}{220, 20, 60}
\definecolor{bus}{RGB}{255, 127, 80}
\definecolor{car}{RGB}{255, 158, 0}
\definecolor{construction}{RGB}{233, 150, 70}
\definecolor{motorcycle}{RGB}{255, 61, 99}
\definecolor{pedestrian}{RGB}{0, 0, 230}
\definecolor{cone}{RGB}{47, 79, 79}
\definecolor{trailer}{RGB}{255, 140, 0}
\definecolor{truck}{RGB}{255, 99, 71}
\definecolor{surface}{RGB}{0, 207, 191}
\definecolor{flat}{RGB}{175, 0, 75}
\definecolor{sidewalk}{RGB}{75, 0, 75}
\definecolor{terrain}{RGB}{112, 180, 60}
\definecolor{manmade}{RGB}{222, 184, 135}
\definecolor{vegetation}{RGB}{0, 175, 0}

\begin{table*}[t]
\centering
\fontsize{10pt}{11pt}\selectfont
\begin{tabular}{l|c|c|c|c|c}
\toprule[1.2pt]
\multicolumn{1}{c|}{\textbf{Method}} & \textbf{Venue}  & \textbf{Image Backbone} & \textbf{Image Size} & \textbf{mIoU (\%)}  & \textbf{Latency (ms)} \\ 
\midrule[0.4pt]
MonoScene\cite{Monoscene}     & CVPR'22         & ResNet-101              & 928 $\times$ 1600    & 6.06                 &          830.1               \\
TPVFormer\cite{TPVFormer}     & CVPR'23         & ResNet-101              & 928 $\times$ 1600    & 27.83                &          320.8              \\
Vampire\cite{Vampire}         & AAAI'24         & ResNet-101              & 256 $\times$ 704    & 28.30                &          349.2               \\ 
CTF-Occ\cite{Occ3d}           & NIPS'23         & ResNet-101              & 928 $\times$ 1600    & 28.53                &            -               \\
SurroundOcc\cite{Surroundocc} & ICCV'23         & ResNet-101              & 800 $\times$ 1333   & 34.40                &          355.6               \\
BEVDet\cite{BEVDet}           & arXiv'21        & ResNet-50               & 256 $\times$ 704    & 19.38                &            -              \\
OccFormer\cite{OccFormer}     & ICCV'23         & ResNet-50               & 928 $\times$ 1600   & 21.93                &          349.2            \\
% BEVFormer\cite{BEVFormer}     & ECCV'22         & ResNet-101              & 928 $\times$ 600    & 26.9                &          -                \\
% FastOcc\cite{FastOcc}& ICRA'24         & ResNet-50               & 320 $\times$ 800    & 34.21                &      \textbf{62.8}$\dagger$ \\
$\text{COTR}^{*}$ \cite{COTR} & CVPR'24         & ResNet-50               & 256 $\times$ 704    & 37.02                &          168.9               \\
FB-Occ\cite{fbocc}            & ICCV'23         & ResNet-50               & 256 $\times$ 704    & \underline{37.39}    &    \underline{129.7}               \\ 
\midrule[0.4pt]
\rowcolor[gray]{0.90}
\multicolumn{1}{c|}{Ours}      & -               & ResNet-50               & 256 $\times$ 704    & \textbf{39.56}       &     \textbf{77.9}          \\
\bottomrule[1.2pt]
\end{tabular}
\caption{Comparison with single-frame methods on the Occ3D-nuScenes validation set. Latency is measured on a single NVIDIA RTX 3090 GPU. The \texttt{"}-\texttt{"} denotes that the associated results are not available. The \texttt{"}$^{*}$\texttt{"} indicates results reproduced using public code.}
\label{single_frame}
\end{table*}

\begin{table}[t!]
\centering
\setlength{\tabcolsep}{2 pt}
\fontsize{9pt}{11pt}\selectfont
\begin{tabular}{l|c|c|c|c}
\toprule[1.2pt]
\multicolumn{1}{c|}{\textbf{Method}} & \textbf{Venue} &\textbf{Image Backbone} & \textbf{Image Size} & \textbf{mIoU  } \\ 
\midrule[0.4pt]
\ BEVFormer  &  ECCV'22 & ResNet-101 & 928$\times$1600  & 26.88 \\
\ FastOcc  &   ICRA'24  & ResNet-101  & 640$\times$1600  & 39.21 \\ 
\ PanoOcc   &  CVPR'24  & ResNet-101 & 864$\times$1600 & 42.13  \\
\ BEVDet4D &   arXiv'21   & ResNet-50  & 384$\times$704 & 39.25\\
\ FB-Occ  &    ICCV'23   & ResNet-50  & 256$\times$704  & 40.69 \\
% \ FastOcc &  ICRA'24  & ResNet-101 & 640$\times$1600 & 40.75 \\
\ COTR   &   CVPR'24    & ResNet-50  & 256$\times$704  & 44.45 \\ 
\rowcolor[gray]{0.90} 
\midrule[0.4pt]
\multicolumn{1}{c|}{Ours} & - & ResNet-50 & 256$\times$704  &  \textbf{45.02} \\ 
\bottomrule[1.2pt] 
\end{tabular}
\caption{Comparison with multi-frame methods on the Occ3D-nuScenes validation set. }
\label{multi_frame}
\end{table}

\subsection{Training}
\subsubsection{Robust Prototype Learning.}
Scene-Adaptive Prototypes $\mathbf{P}^d$ are determined by the class-specific masks $\mathbf{M}^{cls}$ obtained from AdaPG. 
However, when these masks are inaccurately estimated, features from voxels of incorrect classes may be erroneously included in the prototypes $\mathbf{P}^d$, resulting in a decline in overall Occupancy prediction performance.

To address this, RPL injects noise into class-specific masks \(\mathbf{M}^{cls}\) to generate Noisy Scene-Adaptive Prototypes \(\hat{\mathbf{P}}^d\). These prototypes are then combined with the  Scene-Agnostic Prototypes \(\mathbf{P}^g\) to form Noisy Scene-Aware Queries \(\hat{Q}^{SA}\). Subsequently, \(\hat{Q}^{SA}\) is concatenated with the original Scene-Aware Queries \(Q^{SA}\), and these queries are used separately to predict the occupancy and class labels. 

RPL introduces two types of noise to enhance the inference robustness of ProtoOcc: scaling noise and random flipping noise. Scaling noise enlarges or shrinks \(\mathbf{M}^{cls}\) by a random ratio based on the ego vehicle's position, while random flipping noise randomly reallocates voxel grid classes. 
By injecting these perturbations, the model is trained through RPL to effectively denoise and predict occupancy. 
This ensures robust predictions even when the class-specific masks $\mathbf{M}^{cls}$ are inaccurately estimated during inference. 
This approach improves prediction robustness during inference while maintaining computational efficiency, as RPL is applied only during training.

\subsubsection{Training Loss.}
The total loss $\mathcal{L}_{total}$ is given by
\begin{equation}
\mathcal{L}_{total} = \mathcal{L}_{depth} +  \mathcal{L}_{AdaPG} + \mathcal{L}_{occ} + \mathcal{L}_{RPL} ,
\end{equation}
where $\mathcal{L}_{depth}$ is for depth estimation, $\mathcal{L}_{AdaPG}$ is for class-specific mask prediction in AdaPG, $\mathcal{L}_{occ}$ is for query-based occupancy prediction, and $\mathcal{L}_{RPL}$ is for the Robust Prototype Learning. 
Specifically, $\mathcal{L}_{depth}$ employs cross-entropy (CE) loss using LiDAR point clouds projected onto the image.
$\mathcal{L}_{AdaPG}$ includes Lovasz \cite{lovasz} and Dice losses for class-specific mask prediction used in Scene-Adaptive Prototypes generation.  Note that $\mathcal{L}{occ}$ is computed without employing a bipartite matching process, as the prototypes are directly assigned to each class. This loss combines cross-entropy (CE) loss for classification with focal loss \cite{focalloss} and dice loss for mask prediction. Similarly, $\mathcal{L}_{RPL}$ applies the same functions as $\mathcal{L}_{occ}$ to the $\hat{Q}^{SA}$  introduced in RPL.

\section{Experiments}
\subsection{Experimental Settings}

\subsubsection{Dataset and Metrics.}
We conducted the experiments on the Occ3D dataset \cite{Occ3d}, which evaluates the mean Intersection over Union (\textit{mIoU}) across 17 classes. Additionally, we measured the latency of our model.

\subsubsection{Implementation Details.}
We utilized ResNet-50 \cite{ResNet} for the image backbone network.  In DBE, the voxel branch uses a kernel size of 3 for 3D convolution, while the BEV branch employs a kernel size of 7 for 2D convolution.
Our model was trained for 24 epochs with a total batch size of 16 on 4 NVIDIA RTX 3090 GPUs.
The AdamW optimizer was used with a learning rate of $4$$\times$$10^{-4}$ for single-frame and $2$$\times$$10^{-4}$ for multi-frame. 

\subsection{Performance Comparison}
Table \ref{single_frame} presents a detailed comparison of single-frame methods on the Occ3D-nuScenes validation set, demonstrating our method's superior performance. ProtoOcc, utilizing the ResNet-50 backbone, achieves a performance of 39.56\% \textit{mIoU}, outperforming all existing methods \cite{OccFormer, TPVFormer, Surroundocc, Vampire}, including those employing the larger ResNet-101 backbone.
% Notably, our method surpasses the latest state-of-the-art, FB-Occ \cite{fbocc}, by a significant margin of 2.17\% \textit{mIoU}. With an inference time of 77.9 ms, ProtoOcc is slightly slower than FastOcc \cite{FastOcc}, but it achieves a remarkable performance improvement of 5.35\% \textit{mIoU}. These results highlight that ProtoOcc combines high efficiency with superior accuracy, making it well-suited for real-time applications.
% Notably, ProtoOcc achieves the fastest inference time of 77.9ms among existing approaches while demonstrating a remarkable performance improvement of 2.17\% \textit{mIoU} compared to the previous state-of-the-art method. 
Notably, ProtoOcc achieves an inference time of 77.9 ms, demonstrating a 1.7× faster speed compared to the previous state-of-the-art method, while also achieving a remarkable performance improvement of 2.17\% in \textit{mIoU}.
These results demonstrate that ProtoOcc achieves both high efficiency and superior accuracy, making it well-suited for real-time applications.

We also adopt multi-frame methods for ProtoOcc, fusing eight consecutive voxel features over time. Following BEVDet4D \cite{BEVDet}, these voxel features are concatenated along the channel dimension and processed through a residual block followed by a $1\times 1 \times 1$ convolution layer to reduce the channel dimensionality.
Table \ref{multi_frame} provides a performance comparison with other multi-frame methods evaluated on the Occ3D-nuScenes validation set \cite{Occ3d}.
ProtoOcc establishes a new state-of-the-art performance, exhibiting substantial improvements over existing methods \cite{BEVFormer, FastOcc, panoocc, BEVDet, fbocc} and surpassing the previous best model, COTR \cite{COTR}, by 0.57\% in \textit{mIoU}.

\begin{figure*}[ht!]
    \centering
    \includegraphics[scale=0.57]{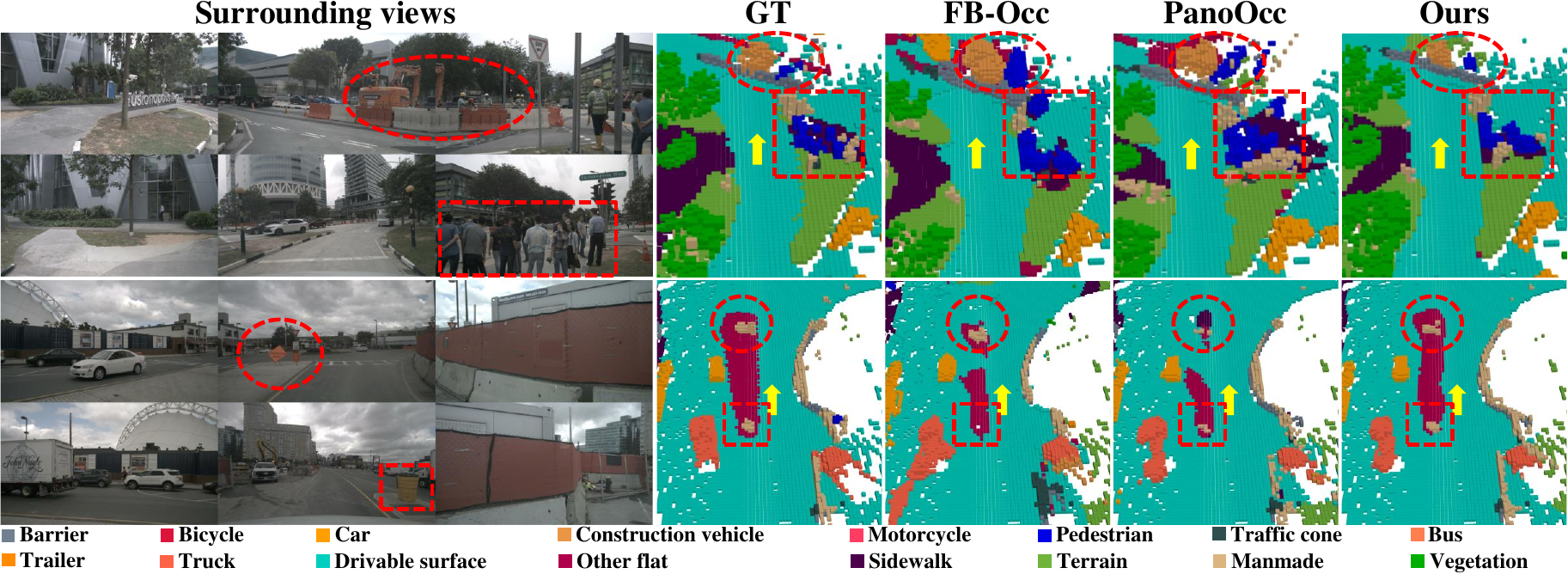}
    \caption{
    Qualitative results on the Occ3D-nuScenes validation set. 
    The regions marked by red ellipses and rectangles emphasize the superior results generated by our proposed model. The yellow arrow indicates the position and direction of the ego vehicle.
    }
    \label{vis1}
\end{figure*}

\subsection{Ablation Study}
We performed an ablation study to evaluate the contributions of the components proposed in ProtoOcc. We trained on a quarter of the dataset for 24 epochs and evaluated the entire validation set using a ResNet-50 backbone \cite{ResNet} with a $256\times704$ resolution and a single frame.

\subsubsection{Contributions of Main Components.}
Table \ref{ablation_main} shows the impact of our main modules. 
The first row denotes a baseline employing 3D CNNs with small kernel sizes for both the encoder and the decoder.
When adding DBE into the baseline, we demonstrate a notable 1.69\% increase in \textit{mIoU}.
This improvement shows that DBE effectively integrates long-range spatial relationships by expanding the receptive field in the BEV domain while capturing fine-grained 3D structures in the voxel domain.
We integrated PQD into the baseline, achieving a 1.45\% \textit{mIoU} improvement while maintaining latency. This demonstrates that PQD effectively captures class distributions through prototypes, enhancing performance without iterative query decoding.
Incorporating both DBE and PQD surpasses the baseline by 2.87\% in \textit{mIoU}.
RPL improves the \textit{mIoU} by an additional 0.4\%, reducing the impact of inaccurate class-specific masks.

\begin{table}[t!]
\centering
\fontsize{9pt}{11pt}\selectfont
\begin{tabular}{c|c|c|>{\hspace{0.05cm}}cc}
\toprule[1.2pt]
\hspace{3pt}\textbf{DBE}\hspace{3pt} & \hspace{3pt}\textbf{PQD}\hspace{3pt} & \hspace{3pt}\textbf{RPL}\hspace{3pt} & \textbf{mIoU} & \textbf{Latency (ms)} \\
\midrule[0.4pt]
\ & & & 34.18 & \textbf{60.0} \\

\
\checkmark & & & 35.87 \textcolor{blue}{(+1.69)} & 75.7   \\
\
 & \checkmark & &35.63 \textcolor{blue}{(+1.45)} & 61.1   \\
\
 \checkmark & \checkmark & & 37.05 \textcolor{blue}{(+2.87)} & 77.9  \\ 
\rowcolor[gray]{0.90} 
\midrule[0.4pt]
 \checkmark & \checkmark & \checkmark & \textbf{37.45} \textcolor{blue}{(\textbf{+3.27})}  &  77.9  \\ \bottomrule[1.2pt] 
\end{tabular}
\caption{Ablation study for evaluating the main components of ProtoOcc.}
\label{ablation_main}
\end{table}

\subsubsection{Contributions of Dual Branch Encoder.}
Table \ref{ablation_dualbranch} presents the results of the ablation study conducted on the Dual Branch Encoder. We focus on the impact of varying kernel sizes within the voxel and BEV branches. We tried kernel sizes of 3 and 7 in the voxel branch, as shown in Table \ref{ablation_dualbranch}. Using a kernel size of 7 in scenario (b) resulted in significant latency increases due to the high dimensionality of 3D space. In contrast, increasing the kernel size within the BEV domain, as demonstrated in scenario (d), led to a comparatively minor latency increase when contrasted with scenario (c). Scenario (b), with its larger voxel branch kernel size, delivered marginally better performance than scenario (d) in the BEV branch.

Further enhancements were observed when the voxel and BEV branches were combined, as seen in scenarios (e) through (g). Specifically, setting the kernel sizes to 3 for the voxel branch and 7 for the BEV branch, and incorporating multi-scale fusion, not only outperformed scenario (b) in terms of performance but also maintained lower latency. The multi-scale fusion alone contributed an increase of 0.23\% in \textit{mIoU} compared to scenario (e), while our specific configuration provided an additional improvement of 0.52\% in \textit{mIoU}.

\subsubsection{Impact of the Prototype Query Decoder.}
Table \ref{ablation_prototype_} presents a comparison of different decoder types, assessing their performance in terms of \textit{mIoU} and latency.
While a query-based decoder \cite{OccFormer} yields higher performance compared to a CNN-based decoder, it incurs much higher latency due to their iterative decoding process. Conversely, when utilizing AdaPG and AgnoPG without iterative decoding, not only do they surpass the query-based decoder by an additional 0.79\% in \textit{mIoU}, but they also achieve a substantial reduction in latency, amounting to 73.7ms.

\begin{table}[t!]
\centering
\fontsize{9pt}{11pt}\selectfont
\setlength{\tabcolsep}{1pt}
\begin{tabular}{c|c|c|c|c|>{\hspace{0.05cm}}cc}
\toprule[1.2pt]
{\multirow{2}{*}{\textbf{Branch}}}& {\multirow{2}{*}{\textbf{Model}}}  & \textbf{Voxel}  & \textbf{BEV}    & \textbf{MS}  & {\multirow{2}{*}{\textbf{mIoU} }} & \textbf{Latency} \\ 
                         &  & \textbf{Kernel} & \textbf{Kernel} & \textbf{Fusion}  &                          & \textbf{(ms)} \\ 
\midrule[0.4pt]
{\multirow{2}{*}{Voxel Only}}& (a) & 3 &  &   & 35.71 & 60.7 \\
\ & (b) & 7 &  &  & 36.14 & 87.2  \\ 
\midrule[0.4pt]
\ {\multirow{2}{*}{BEV Only}} & (c) &  & 3 &  & 35.64 & \textbf{52.0} \\
\ & (d) &  & 7 &  & 36.07 & 55.1 \\ 
\midrule[0.4pt]
\  & (e) & 3 & 3 &   & 36.70 & 71.1 \\ 
\  & (f) & 3 & 3 &    \checkmark  & 36.93 & 74.5 \\ 
\ {\multirow{-3}{*}{Dual Branch}} &\cellcolor[gray]{0.90} Ours, (g)  \hspace{0.5pt} & \cellcolor[gray]{0.90}{3} & \cellcolor[gray]{0.90}7 & \cellcolor[gray]{0.90}\hspace{1pt}\checkmark & \cellcolor[gray]{0.90}\textbf{37.45} & \cellcolor[gray]{0.90}77.9 \\ 
\bottomrule[1.2pt]
\end{tabular}
\caption{Ablation study for Dual Branch Encoder.
 \textit{MS Fusion} indicates the use of multi-scale fusion in HFM.
}
\label{ablation_dualbranch}
\end{table}

\subsection{Qualitative Analysis}
Figure \ref{vis1} presents qualitative results on the Occ3D-nuScenes validation set, comparing the proposed model with FB-Occ \cite{fbocc} and PanoOcc \cite{panoocc}. 
ProtoOcc provides accurate predictions in complex scenes, particularly for regions with ambiguous boundaries and diverse object types. 

\begin{table}[t!]
\centering
\fontsize{9pt}{11pt}\selectfont
\setlength{\tabcolsep}{1.8pt}
\begin{tabular}{c|c|c|c|>{\hspace{0.05cm}}c>{\hspace{0.05cm}}c}
\toprule[1.2pt]
{\multirow{2}{*}{\textbf{Decoder Type}}} & \multirow{2}{*}{\textbf{AdaPG}} & \multirow{2}{*}{\textbf{AgnoPG}} & \textbf{Iterative}  & {\multirow{2}{*}{\textbf{mIoU}}} & \textbf{Latency}   \\  
 & & & \textbf{Decoding} & & \textbf{(ms)}\\
\midrule[0.4pt]  
\  CNN-based &  &  &  & 35.87 & \hspace{0.5pt} \textbf{76.1} \\ 
\midrule[0.4pt] 
% \ {\multirow{2}{*}{Query}}   & L & -  & 33.25 & \textbf{61.6} \\
% \    & L  & \checkmark & 36.66 & 151.6 \vspace*{3pt} \\ 
\ Query-based &  &  & \checkmark  & 36.66 & 151.6 \\ 
\midrule[0.4pt] % \vspace*{3pt} \hline  
\    &  \hspace{0.5pt} \checkmark &  &  & 36.87 & 77.4 \\ 
% \ PQD   &  & \checkmark &  & 36.74 & \textbf{61.6} \\ 
\rowcolor[gray]{0.90} 
\ \cellcolor{white} {\multirow{-2}{*}{PQD}} & \cellcolor[gray]{0.90} \checkmark & \cellcolor[gray]{0.90}\checkmark &\cellcolor[gray]{0.90}  & \cellcolor[gray]{0.90}\textbf{37.45} & \cellcolor[gray]{0.90}77.9 \\ 
\bottomrule[1.2pt] 
\end{tabular}
\caption{Comparison of different decoder types. 
}
\label{ablation_prototype_}
\end{table}

\section{Conclusions}
In this paper, we introduced ProtoOcc, an efficient encoder-decoder framework designed for 3D occupancy prediction. The DBE leverages both voxel and BEV representations, capturing fine-grained interactions and efficiently modeling long-range spatial relationships to enhance encoder performance. Furthermore, the PQD employs Scene-Adaptive and Scene-Agnostic Prototypes as queries, which eliminate the need for an iterative decoding process, thereby significantly reducing computational complexity. We also introduced the RPL to increase the model’s robustness against inaccuracies in prototypes. Our method achieved state-of-the-art performance with faster inference speeds on the Occ3D-nuScenes benchmark.

\bibliography{protoocc}

\clearpage
%DY
\appendix % Supplementary section 시작
% \section*{Supplementary Materials for ProtoOcc}
\twocolumn[
\begin{center}
    {\LARGE \textbf{Supplementary Materials for ProtoOcc}}
    \vspace{50pt}
\end{center}
]

\begin{table}[h]
\begin{minipage}{\textwidth}
\vspace{-20pt}
\centering
\setlength{\tabcolsep}{1.6pt}
\fontsize{8pt}{9pt}\selectfont
\begin{tabular}{l|c|c|ccccccccccccccccc|c}
\toprule[1.2pt]
\textbf{Method} & Backbone & Resolution &
  \rotatebox{90}{\coloredsquare{others} \textbf{others}} &
  \rotatebox{90}{\coloredsquare{barrier} \textbf{barrier}} &
  \rotatebox{90}{\coloredsquare{bicycle} \textbf{bicycle}} &
  \rotatebox{90}{\coloredsquare{bus} \textbf{bus}} &
  \rotatebox{90}{\coloredsquare{car} \textbf{car}} &
  \rotatebox{90}{\coloredsquare{construction} \textbf{construction}} &
  \rotatebox{90}{\coloredsquare{motorcycle} \textbf{motorcycle}} &
  \rotatebox{90}{\coloredsquare{pedestrian} \textbf{pedestrian}} &
  \rotatebox{90}{\coloredsquare{cone} \textbf{traffic cone}} &
  \rotatebox{90}{\coloredsquare{trailer} \textbf{trailer}} &
  \rotatebox{90}{\coloredsquare{truck} \textbf{truck}} &
  \rotatebox{90}{\coloredsquare{surface} \textbf{driveable surface}} &
  \rotatebox{90}{\coloredsquare{flat} \textbf{other flat}} &
  \rotatebox{90}{\coloredsquare{sidewalk} \textbf{sidewalk}} &
  \rotatebox{90}{\coloredsquare{terrain} \textbf{terrain}} &
  \rotatebox{90}{\coloredsquare{manmade} \textbf{manmade}} &
  \rotatebox{90}{\coloredsquare{vegetation} \textbf{vegetation}} &  
  \rotatebox{90}{\textbf{mIoU (\%)}} \\ \midrule[0.4pt]
MonoScene     & R101  & 600$\times$928  & 1.75  & 7.23  & 4.26  & 4.93  & 9.38  & 5.67  & 3.98  & 3.01  & 5.90  & 4.45  & 7.17  & 14.91 & 6.32  & 7.92  & 7.43  & 1.01  & 7.65  & 6.06   \\
OccFormer     & R101  & 928$\times$1600 & 5.94  & 30.29 & 12.31 & 34.40 & 39.17 & 14.44 & 16.45 & 17.22 & 9.27  & 13.9  & 26.36 & 50.99 & 30.96 & 34.66 & 22.73 & 6.76  & 6.97  & 21.93  \\
TPVFormer     & R101  & 600$\times$928  & 7.22  & 38.90 & 13.67 & 40.78 & 45.9  & 17.23 & 19.99 & 18.85 & 14.30 & 26.69 & 34.17 & 55.65 & 35.47 & 37.55 & 30.70 & 19.40 & 16.78 & 27.83  \\
CTF-Occ       & R101  & 640$\times$960  & 8.09  & 39.33 & 20.56 & 38.29 & 42.24 & 16.93 & 24.52 & 22.72 & 21.05 & 22.98 & 31.11 & 53.33 & 33.84 & 37.98 & 33.23 & 20.79 & 18.00 & 28.53  \\
SurroundOcc   & R101  & 800$\times$1333 & 9.51  & 38.50 & 22.08 & 39.82 & 47.04 & 20.45 & 22.48 & 23.78 & 23.00 & 27.29 & 34.27 & 78.32 & 36.99 & 46.27 & 49.71 & 35.93 & 32.06 & 34.60  \\
BEVDet        & R50   & 256$\times$704  & 4.39  & 30.31 & 0.23  & 32.36 & 34.47 & 12.97 & 10.34 & 10.36 & 6.26  & 8.93  & 23.65 & 52.27 & 24.61 & 26.06 & 22.31 & 15.04 & 15.10 & 19.38  \\
Vampire       & R50   & 256$\times$704  & 7.48  & 32.64 & 16.15 & 36.73 & 41.44 & 16.59 & 20.64 & 16.55 & 15.09 & 21.02 & 28.47 & 67.96 & 33.73 & 41.61 & 40.76 & 24.53 & 20.26 & 28.33  \\
% FastOcc       & R50   & 320$\times$800 & - & - & - & - & - & - &- & - & - & - & - & - & - & - & - & - & - & 34.21\\ 
COTR$^{*}$         & R50   & 256$\times$704  & 9.44  & 42.47 & 23.03 & \underline{43.04} & \underline{49.23} & \underline{23.45} & 24.21 & \underline{26.28} & 25.66 & 27.78 & 34.77 & \underline{79.81} & \underline{41.97} & \underline{49.86} & \underline{52.77} & \underline{40.49} & \underline{35.08} & 37.02  \\
FB-Occ        & R50   & 256$\times$704 & \underline{12.17} & \underline{44.83} &\underline{25.73} & 42.61 & 47.97 & 23.16 & \underline{25.17} & 25.77 & \underline{26.72} & \underline{31.31} & \underline{34.89} & 78.83 & 41.42 & 49.06 & 52.22 & 39.07 & 34.61 & \underline{37.39}  \\ 
% \noalign{\vskip 0.03in} \hline \noalign{\vskip 0.03in}
\midrule[0.4pt]
\rowcolor[gray]{0.90} 
 % Ours & R50   & 256$\times$704  & \textbf{12.40} & \textbf{46.97} & \underline{25.41} &\textbf{45.41} & \textbf{51.94} & \textbf{24.53} & \textbf{26.69} & \textbf{27.63}& \textbf{27.46} & \textbf{32.95} & \textbf{36.40} & \textbf{81.77} & \textbf{45.06} & \textbf{52.60} & \textbf{55.89} & \textbf{42.14} & \textbf{36.59} & \textcolor{red}{\textbf{39.52}}  \\
 Ours & R50   & 256$\times$704  & \textbf{12.39} & \textbf{45.94} & \textbf{26.27} &\textbf{44.40} & \textbf{51.78} & \textbf{26.80} & \textbf{27.57} & \textbf{27.97}& \textbf{27.46} & \textbf{32.79} & \textbf{36.89} & \textbf{81.82} & \textbf{45.71} & \textbf{53.06} & \textbf{56.48} & \textbf{42.15} & \textbf{36.57} & \textbf{39.56} \\
\bottomrule[1.2pt] 
\end{tabular}
\vspace{-5pt}
% \caption{\textcolor{red}{Class-wise performance comparison} with previous single-frame methods on the Occ3D-nuScenes validation Set. The \texttt{"}R50\texttt{"} and \texttt{"}R101\texttt{"} respectively correspond to ResNet-50 \cite{ResNet} and ResNet-101. Latency is measured on a single NVIDIA RTX 3090 GPU. \textcolor{red}{The \texttt{"}$\dagger$\texttt{"} denotes that the latency was measured on an NVIDIA V100 GPU as reported in the paper.} The \texttt{"}$^{*}$\texttt{"} indicates results reproduced using public code.
\caption{Comparison of class-wise performance with previous single-frame methods on the Occ3D-nuScenes validation set. The \texttt{"}R50\texttt{"} and \texttt{"}R101\texttt{"} respectively correspond to ResNet-50 \cite{ResNet} and ResNet-101. The \texttt{"}$^{*}$\texttt{"} indicates results reproduced using public code. The \texttt{"}-\texttt{"} denotes that the associated results are not available.
}
\label{tab:per_cls_single_frame_performance}
\end{minipage}
\end{table}

% \vspace{-5pt}

% We provide additional implementation details, experiment results, and visualization analysis for our proposed ProtoOcc. 

In this Supplementary Materials, we present more details that could not be included in the main paper due to space limitations.
We discuss the following:
\begin{itemize}
    \setlength\itemsep{0.2em}
    \setlength\parsep{0.2em}  
    \setlength\parskip{0.2em}
    \item Further experiments details;
    \item Additional experiment results for occupancy prediction;
    \item Extensive qualitative analysis of ProtoOcc.
\end{itemize}

\section{Further Experiments Details}
\subsection{Datasets}
The nuScenes dataset \cite{nuScenes} consists of 700 training scenes and 150 validation scenes, with a duration of 20 seconds per scene. Key samples in each scene are annotated at a 2 Hz frequency, resulting in 28,130 training samples and 6,019 validation samples. 
The Occ3D-nuScenes dataset is designed for occupancy prediction from the nuScenes dataset.
The semantic occupancy ground truth (GT) covers a range of [-$40m$, -$40m$, -$1m$, $40m$, $40m$, $5.4m$] with a voxel size of [$0.4m$, $0.4m$, $0.4m$] in the ego coordinate system. 
Additionally, visibility masks for both LiDAR and camera modalities are provided, enabling the evaluation of model performance in areas visible to each sensor. While the voxels are categorized into 18 classes, including the "empty" category, the evaluation focuses on 17 semantic classes without "empty".

\section*{ }
\vspace{198pt}

\begin{table}[h!]
\centering
\setlength{\tabcolsep}{2 pt}
\fontsize{9pt}{9pt}\selectfont

\begin{tabular}{l|c|c|c|c}
\toprule[1.2pt]
\multicolumn{1}{c|}{\textbf{Method}} & \textbf{Venue} &\textbf{Backbone} & \textbf{mIoU  } & \textbf{Latency (ms)}\\ 
\midrule[0.4pt]
LMSCNet*\cite{lmscnet}               & 3DV'20          & EB7  & 6.70                   &   -               \\
3DSketch*\cite{3dsketch}             & CVPR'20         & EB7 & 7.50                 &   -               \\
AICNet*\cite{aicnet}                 & CVPR'20         & EB7 & 8.31                   &   -               \\
JS3C-Net*\cite{js3c-net}             & AAAI'21         & EB7 & 10.31                &   -               \\
MonoScene\cite{Monoscene}            & CVPR'22         & EB7    & 11.08                &   478.87               \\
TPVFormer\cite{TPVFormer}            & CVPR'23         & EB7    & 11.36                &   323.03              \\
OccFormer\cite{OccFormer}            & ICCV'23         & EB7    & \underline{13.46}    &   377.13            \\
SparseOcc\cite{SparseOcc_Rethinking} & CVPR'24         & EB7 & 13.12                &   \underline{240.35}               \\ 
\midrule[0.4pt]
\rowcolor[gray]{0.90}
\multicolumn{1}{c|}{Ours}      & -               & EB7         & \textbf{13.89}       &     \textbf{71.88}          \\
\bottomrule[1.2pt] 
\end{tabular}
\caption{Comparison with single-frame methods on the SemanticKITTI \cite{semantickitti} validation set. The methods with \texttt{"}*\texttt{"} are RGB-input variants reported by \cite{Monoscene} for a fair comparison.
The \texttt{"}EB7\texttt{"} correspond to EfficientNet-B7 \cite{EfficientNet}. Latency is measured on a single NVIDIA RTX 3090 GPU.
}
\label{semantic_kitti_single_frame}
\end{table}

% \vspace{-10pt}

\subsection{Metrics}
In occupancy prediction tasks, \textit{mean Intersection over Union} (\textit{mIoU}) is utilized to evaluate the model performance. This metric measures the overlap between the predicted class and the GT class for each voxel and then averages this over all semantic classes.
The metric is defined as follows:
\begin{equation}
\text{mIoU} = \frac{1}{C} \sum_{c=1}^{C} \frac{\text{TP}_c}{\text{TP}_c + \text{FP}_c + \text{FN}_c},
\end{equation}
where \(C\) represents the number of semantic categories. The \(\text{TP}_c\), \(\text{FP}_c\), and \(\text{FN}_c\) correspond to the number of true positives, false positives, and false negatives, respectively.
% 3. mIoU metric

\subsection{Implementation Details}
ProtoOcc adopts ResNet-50 \cite{ResNet} as the image backbone with an input resolution of $256\times704$. We apply data augmentation methods to the input images, such as random flipping, rotation, and resizing crops. After the 2D-to-3D view transformation, only random flipping along the X and Y axes is applied in the voxel domain.
In DBE, a small kernel size of 3 was applied to 3D CNNs in the Voxel Branch, and a large kernel size of 7 was used for 2D CNNs in the BEV Branch.
The EMA coefficient $\alpha$ in AgnoPG was set to 0.01.
Our model was trained on a system running Ubuntu 18.04, equipped with two Intel Xeon CPUs and four 24G NVIDIA RTX 3090 GPUs. The training was conducted for 24 epochs with a total batch size of 16.
The AdamW optimizer was employed with a learning rate of $4$$\times$$10^{-4}$ and a weight decay of 0.01. 
For the multi-frame setting, we used 8 frames and adjusted the learning rate to $2$$\times$$10^{-4}$, while maintaining all other parameters the same as in the single-frame setup.

\begin{table}[!t]
\centering
\fontsize{9pt}{10pt}\selectfont
\setlength{\tabcolsep}{5.0pt}
\begin{tabular}{c|c|c|cc|cc}
\toprule[1.2pt]
{\multirow{2}{*}{\textbf{Branch}}} & \textbf{BEV} & \textbf{Voxel} & \multicolumn{2}{c|}{\multirow{2}{*}{\textbf{mIoU}}}  & \multicolumn{2}{c}{\textbf{Latency}} \\ 
 & \textbf{Kernel} & \textbf{Kernel} & & & \multicolumn{2}{c}{\textbf{(ms)}} \\ 
\midrule[0.4pt]
\multirow{5}{*}{\shortstack{BEV\\Branch}} & 3 & -  & 35.79 & - & \textbf{52.0} & - \\
% \multirow{4}{*}{BEV\\Branch} & 3 & -  & 35.79 & - & \textbf{52.0} & - \\
 & 5 & -    & 36.20 & \textcolor{blue}{+0.41} & 53.9 & \textcolor{red}{+1.9} \\ 
 & 7 & -    & 36.37 & \textcolor{blue}{+0.58} & 55.1 & \textcolor{red}{+3.1} \\
 & 9 & -    & 36.26 & \textcolor{blue}{+0.47} & 55.8 & \textcolor{red}{+3.8} \\
 & 11 & -   & 36.38 & \textcolor{blue}{+0.59} & 56.4 & \textcolor{red}{+4.4} \\ 
\midrule[0.4pt]
\multirow{5}{*}{\shortstack{Voxel\\Branch}} & - & 3 & 35.71 & - & 60.7 & - \\
% \multirow{5}{*}{Voxel\\Branch} & - & 3 & 35.71 & - & 60.7 & - \\
 & - & 5    & 35.87 & \textcolor{blue}{+0.16} & 80.4 & \textcolor{red}{+19.7} \\ 
 & - & 7    & 36.14 & \textcolor{blue}{+0.43} & 87.2 & \textcolor{red}{+26.5} \\ 
 & - & 9    & 36.34 & \textcolor{blue}{+0.63} & 98.6 & \textcolor{red}{+37.9} \\ 
 & - & 11   & 36.19 & \textcolor{blue}{+0.48} & 118.5 & \textcolor{red}{+57.8} \\ 
\midrule[0.4pt]
\multirow{5}{*}{\shortstack{Dual\\Branch}} & 3 & 3   & 36.93 & - & 74.5 & - \\
% \multirow{5}{*}{Dual\\Branch} & 3 & 3   & 36.93 & - & 74.5 & - \\
 & 5 & 3   & 37.09 & \textcolor{blue}{+0.16} & 77.6 & \textcolor{red}{+3.1} \\ 
 &\cellcolor[gray]{0.90}7 &\cellcolor[gray]{0.90}3   &\cellcolor[gray]{0.90}\textbf{37.45} & \cellcolor[gray]{0.90}\textcolor{blue}{+0.52} &\cellcolor[gray]{0.90}77.9 & \cellcolor[gray]{0.90}\textcolor{red}{+3.4} \\
 & 9 & 3   & 37.41 & \textcolor{blue}{+0.48} & 78.1 & \textcolor{red}{+3.6} \\
 & 11 & 3  & \textbf{37.45} & \textcolor{blue}{+0.52} & 78.2 & \textcolor{red}{+3.7} \\ 
\bottomrule[1.2pt]
\end{tabular}
\caption{
Ablation study on kernel sizes for Voxel, BEV, and Dual branches. The blue numbers indicate changes in \textit{mIoU}, while the red numbers represent changes in latency, both relative to the baseline (first row) of each section.
}
\label{ablation_kernelsize}
\end{table}

\vspace{5pt}
\section{Additional Experimental Results}

This section presents class-wise performance comparisons with existing methods and additional ablation studies.

\vspace{5pt}
\subsection{Class-wise Performance Comparisons} % with Previous Methods} 
As shown in Table \ref{tab:per_cls_single_frame_performance}, we evaluated the class-wise performance of ProtoOcc against other single-frame methods, including SOTA \cite{Monoscene, OccFormer, TPVFormer, Occ3d, Surroundocc, BEVDet, Vampire, FastOcc, COTR, fbocc} on the Occ3D-nuScenes \cite{Occ3d} validation set.
Notably, ProtoOcc outperforms existing methods in all individual classes, demonstrating its enhanced generalization capabilities to predict occupancy and semantics accurately.
% Notably, ProtoOcc outperforms all previous methods in the majority of individual classes, demonstrating its enhanced generalization capabilities to predict occupancy and semantics accurately.

\subsection{Results on SemanticKITTI.}
% We further investigate the generalizability of ProtoOcc. As depicted in Table \cite{SemanticKITTI}, our method achieves a notable performance of 13. 89\% \textit{mIoU}, surpassing previous methods \cite{Monoscene, TPVFormer, OccFormer, SparseOcc_Rethinking}. 
% In particular, ProtoOcc demonstrates high inference efficiency. ProtoOcc runs at 71.88 ms, which is approximately 3× faster than SparseOcc (240.35 ms) and over 5× faster than OccFormer (377.13 ms). 
To evaluate the generalizability of ProtoOcc, we conducted additional experiments on the SemanticKITTI \cite{semantickitti} validation set. Table \ref{semantic_kitti_single_frame} shows that ProtoOcc outperforms other methods, achieving 13.89\% \textit{mIoU}.
Significantly, ProtoOcc achieves an inference time of 71.88 ms, approximately 3× faster than SparseOcc (240.35 ms) and over 5× faster than OccFormer (377.13 ms).
% As depicted in Table \ref{single_frame} and \ref{semantic_kitti_single_frame}, ProtoOcc achieves high performance on both the SemanticKITTI and Occ3D-nuScenes datasets, underscoring its generalization capabilities.
% he high performance in both SemanticKITTI and Occ3D-nuScenes \cite{Occ3d} demonstrates the generalization capabilities of ProtoOcc.

\vspace{5pt}
\subsection{Additional Ablation Study}
% We conducted additional ablation studies with the same settings as mentioned in the main paper.
For the ablation studies, we trained our model on 1/4 of the Occ3D-nuScenes training dataset and performed evaluations on the full validation set.

\vspace{5pt}
\subsubsection{Comparison of Kernel Size in DBE.}
Table \ref{ablation_kernelsize} presents the results of varying kernel sizes in the BEV, Voxel, and Dual branches. 
Expanding the kernel size in the BEV and Voxel Branches enhances \textit{mIoU}, respectively.
When the kernel size was increased, we observed a significant increase in latency in the Voxel Branch compared to the BEV Branch. 
Based on these results, we focus on expanding the kernel size in the BEV Branch of DBE.
% To achieve a balance between performance and efficiency, we focused on expanding the kernel size in the BEV Branch of DBE. 
The results in the Dual Branch demonstrate the importance of capturing local details in the voxel domain alongside comprehensive context in the BEV domain for accurate occupancy prediction.

\begin{table}[!t]   
\centering
% \fontsize{9pt}{9pt}\selectfont
\fontsize{9pt}{11pt}\selectfont
\setlength{\tabcolsep}{12pt}
\begin{tabular}{c|c|c|c}
\toprule[1.2pt]
\textbf{Pooling Method} & \textit{Sum} & \textit{Max} & \textit{Average} \\
\midrule[0.4pt]
% \  &  &   & 37.05 & \hspace*{0.2cm} \\
\ \textbf{mIoU} & 37.15 & 37.35 & \textbf{37.45}  \\ \bottomrule[1.2pt]
\end{tabular}
% \caption{Ablation study of the operation for prototype generation in AdaPG.}
\caption{Ablation study of the pooling method for prototype generation in AdaPG.}
\label{ablation_AdaPG}
\end{table}

\begin{table}[!t]
\centering
% \fontsize{9pt}{9pt}\selectfont
\fontsize{9pt}{11pt}\selectfont
\setlength{\tabcolsep}{9pt}
\begin{tabular}{c|c|c|c|c}
\toprule[1.2pt]
\textbf{Coefficient $\alpha$} & 0.1 & 0.01 & 0.001 & 0.0001 \\
\midrule[0.4pt]
% \  &  &   & 37.05 & \hspace*{0.2cm} \\
\ \textbf{mIoU} & 37.35 & \textbf{37.45} & 37.33 & 37.28  \\ \bottomrule[1.2pt]
\end{tabular}
\caption{Ablation study of coefficient of EMA in AgnoPG.}
\label{ablation_AgnoPG}
\end{table}

\begin{table}[!t]
\centering
\fontsize{9pt}{11pt}\selectfont
\setlength{\tabcolsep}{10.5pt}
\begin{tabular}{c|c|c|c}
\toprule[1.2pt]
% \textbf{Additional Query} & \textbf{Point Noise} & \textbf{Scale Noise} & \textbf{mIoU} \\
\textbf{Point Noise} & \textbf{Scale Noise} & \textbf{mIoU} & \textbf{Latency} \\
\midrule[0.4pt]
\   &   & 37.05 & 77.9 \\
\ \checkmark &  & 37.25 & 77.9\\
\  & \checkmark  & 37.35 & 77.9 \\
\rowcolor[gray]{0.90}
\  \checkmark & \checkmark & \textbf{37.45} & 77.9\\ \bottomrule[1.2pt]
\end{tabular}
\caption{Ablation study of noise type in RPL.}
% \caption{Ablation study of Robust Prototype Learning.}
\label{ablation_denoise}
\end{table}

\begin{figure*}[!t]
    \centering
    \includegraphics[scale=0.53]{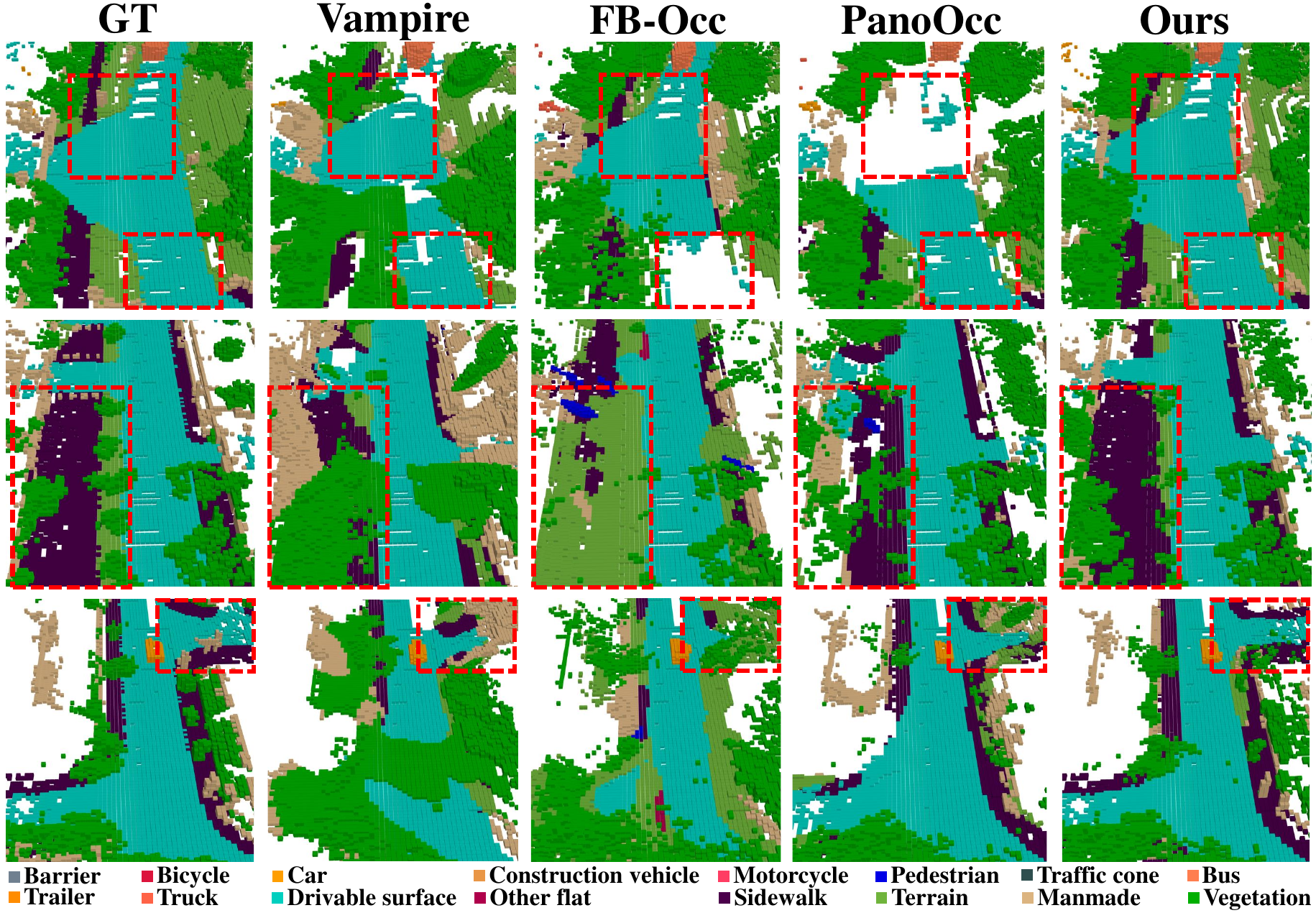}
    \caption{
    Comparison with SOTA methods using qualitative visualization under different scenarios on the Occ3D-nuScenes validation set. The regions marked by red rectangles emphasize the superior results generated by our proposed model. From the first to the fifth column: Ground truth, Vampire, FB-Occ, PanoOcc, and Ours.
    }
    \label{compare_others}
\end{figure*}

\subsubsection{Effect of Prototype Generation Method in AdaPG.} % (CW)
Table \ref{ablation_AdaPG} presents the performance of various pooling methods for prototype generation in AdaPG. ProtoOcc achieves the best performance when \textit{Average pooling} is utilized. This indicates that \textit{Average pooling} provides more balanced features for each class, leading to superior performance in PQD.

\subsubsection{Effect of EMA Coefficient in AgnoPG.} 
As shown in Table \ref{ablation_AgnoPG}, we investigated the impact of the EMA coefficient $\alpha$ in AgnoPG.
ProtoOcc achieved the best performance when $\alpha$ is 0.01.

\subsubsection{Impact of RPL.}
Table \ref{ablation_denoise} shows the effect of RPL, and the first row denotes the baseline. 
When merging Scene-Aware Queries $Q^{SA}$ with Noisy Scene-Aware Queries $\hat{Q}^{SA}$ augmented by point noise and then training the model to predict occupancy for each, we observed a 0.20\% \textit{mIoU} performance improvement. The application of scale noise results in a 0.30\% \textit{mIoU} gain over the baseline. When both noise types are applied simultaneously, the model achieves a 0.40\% \textit{mIoU} improvement over the baseline.
These results indicate that each type of noise is effective for training in occupancy prediction.
By leveraging RPL only during training, the model enhances both robustness and accuracy in occupancy prediction without increasing latency.

\section{Extensive Qualitative Analysis} % (DY)
In this section, we present additional qualitative results of ProtoOcc. We provide visualizations of further comparisons with other models, various weather conditions, and the types of noise used in RPL.

\subsection{Qualitative Results under Different Scenarios} % FB-Occ, PanoOcc, 
We present additional qualitative results with the previous methods, as shown in Figure \ref{compare_others}. 
Our model consistently outperforms existing methods across a wide range of scenarios. 

% \newpage
\subsection{Qualitative Results under Various Weather}
We present qualitative results under various weather conditions in Figures \ref{vis_weather1} and \ref{vis_weather2}. These results demonstrate that ProtoOcc maintains stable and robust occupancy prediction quality, even under more challenging conditions such as rain or night.

% \newpage
\subsection{Visualization of Noise Types in RPL}
Figure \ref{vis_noise} illustrates the types of noise applied in RPL. The additional prototypes with noise are generated only during the training phase. 

By training the model to counteract these noises, we improve its ability to accurately predict semantic occupancy even with low-quality prototypes.

\begin{figure*}[!t]
    \centering
    \includegraphics[scale=0.57]{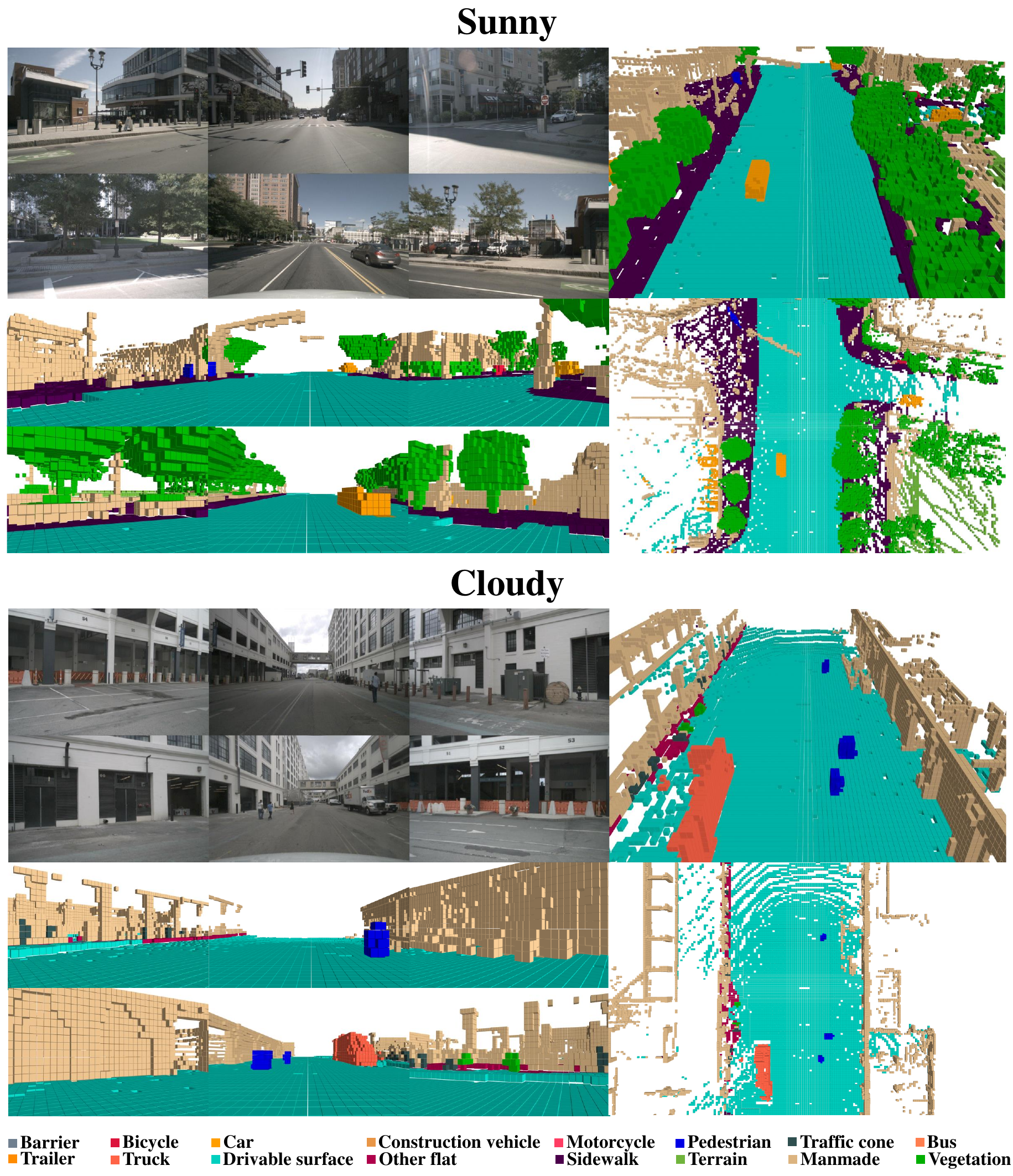}
    \caption{
    Qualitative results under sunny and cloudy conditions on Occ3D-nuScenes validation set.
    }
    \label{vis_weather1}
\end{figure*}

\begin{figure*}[!t]
    \centering
    \includegraphics[scale=0.57]{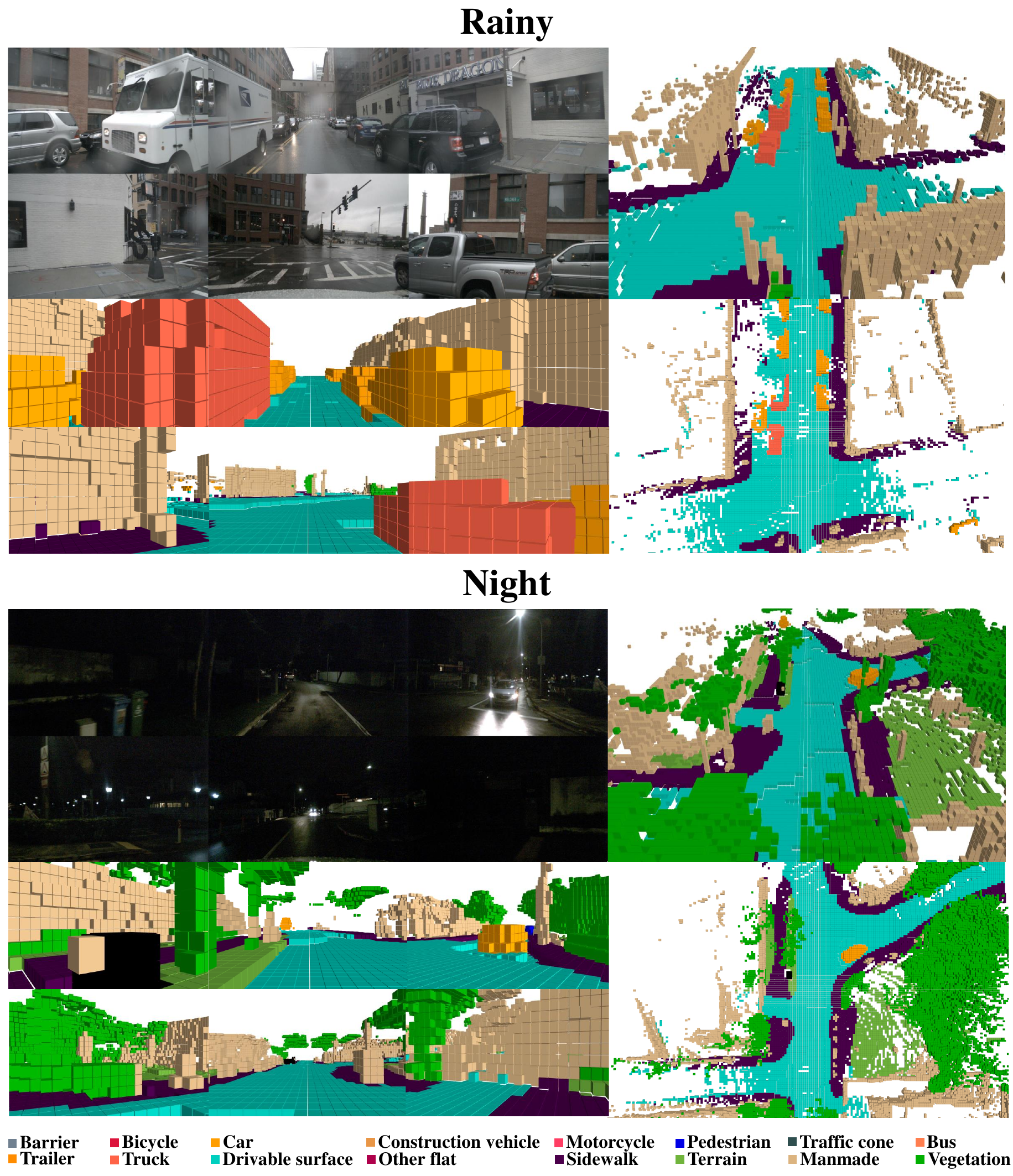}
    \caption{
    Qualitative results under rainy and night conditions on Occ3D-nuScenes validation set.
    }
    \label{vis_weather2}
\end{figure*}

\begin{figure*}[t!]
    \centering
    \includegraphics[scale=0.575]{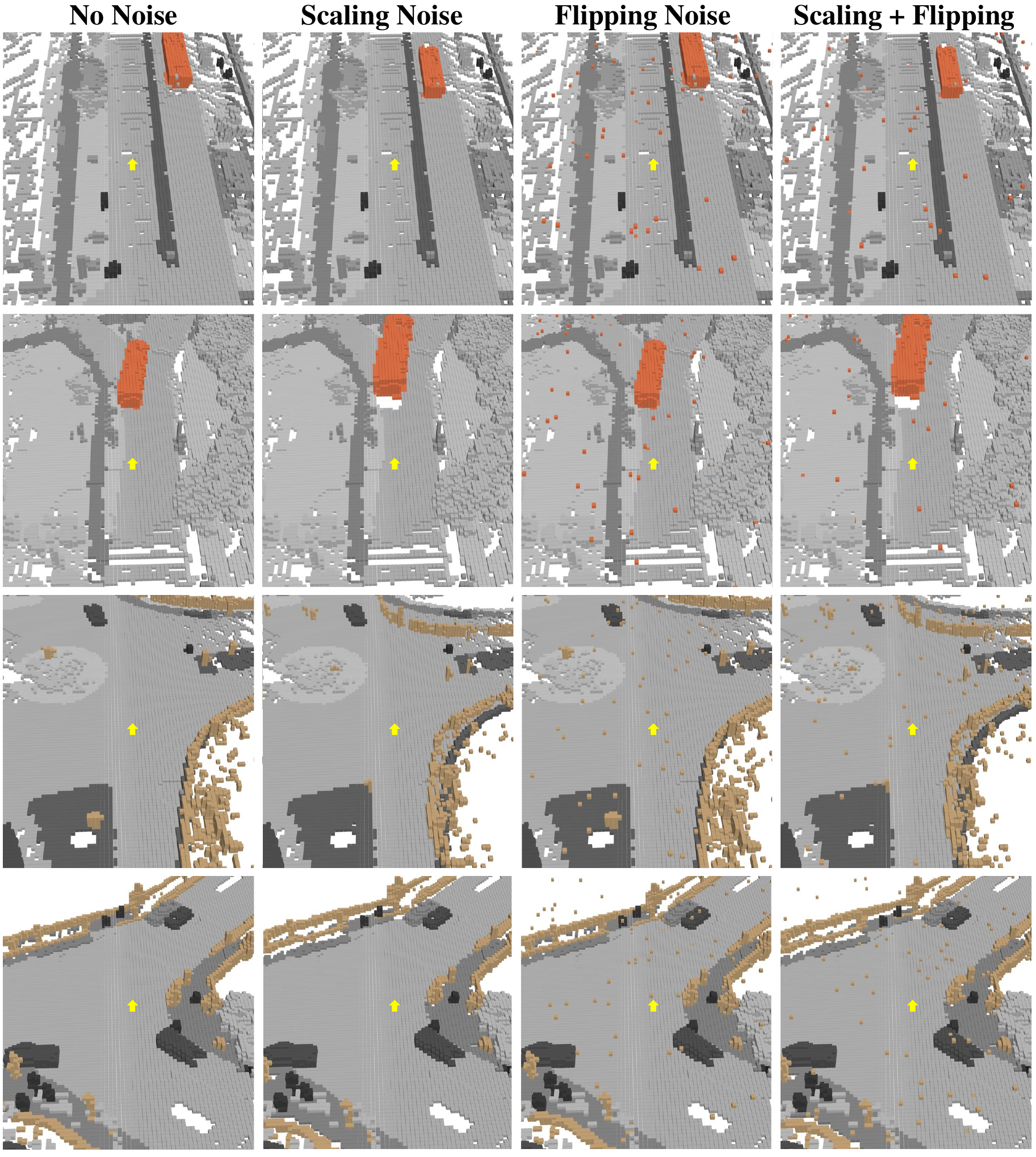}
    \caption{
    Visualization of noise types in RPL. 
    To demonstrate the concept of noise types, we visualize class-specific masks of GT.
    The yellow arrows indicate the position and direction of the ego vehicle. Orange objects represent buses, while caramel-colored areas indicate manmade. Scaling noise adjusts the size of class-specific masks based on the ego vehicle's position. Random flipping noise reallocates voxel grid classes randomly.
    % (DY) noisy 라는 단어의 중복 During the training phase, RPL uses these noisy masks to generate corrupted prototypes, enhancing the model's robustness for accurate predictions even when class-specific masks are incomplete.
    During the training phase, RPL generates noisy prototypes using the predicted class-specific masks with noise.
    %, which enhances the model's robustness for accurate predictions even when class-specific masks are incomplete.
}
    \label{vis_noise}
\end{figure*}

% \newpage
% \pagebreak

\end{document}